\documentclass[10pt,twocolumn,letterpaper]{article}
\pdfoutput=1

\usepackage{iccv}
\usepackage{times}
\usepackage{epsfig}
\usepackage{graphicx}
\usepackage{amsmath}
\usepackage{amssymb}
\usepackage{booktabs}
\usepackage{multirow}

\usepackage[breaklinks=true,bookmarks=false,linkcolor=blue,filecolor=blue,citecolor=black,urlcolor=cyan]{hyperref}
 \hypersetup{
     colorlinks=true,
     linkcolor=blue,
     filecolor=blue,
     citecolor=black,
     urlcolor=cyan,
     }

\iccvfinalcopy 

\def\iccvPaperID{****} 

\usepackage{cleveref}
\crefname{section}{§}{§§}
\Crefname{section}{§}{§§}

\ificcvfinal\fi

\begin{document}

\title{Prior-Enhanced Few-Shot Segmentation with Meta-Prototypes}

\author{Jian-Wei Zhang$^1$ ~~~~ Lei Lv$^1$ ~~~~ Yawei Luo$^1$ ~~~~ Hao-Zhe Feng$^1$ ~~~~ Yi Yang$^2$ ~~~~ Wei Chen$^{1}$\thanks{Corresponding author} \\
$^1$Zhejiang University \\ 
$^2$University of Technology Sydney \\
{\tt\small \{zjw.cs, lvlei, fenghz\}@zju.edu.cn  yaweiluo329@gmail.com} \\ 
{\tt\small Yi.Yang@uts.edu.au  chenwei@cad.zju.edu.cn} 
}

\maketitle
\ificcvfinal\thispagestyle{empty}\fi

\begin{abstract}
    Few-shot segmentation~(FSS) performance has been extensively promoted by introducing episodic training and class-wise prototypes. However, the FSS problem remains challenging due to three limitations: (1) Models are distracted by task-unrelated information; (2) The representation ability of a single prototype is limited; (3) Class-related prototypes ignore the prior knowledge of base classes. We propose the Prior-Enhanced network with Meta-Prototypes to tackle these limitations. The prior-enhanced network leverages the support and query (pseudo-) labels in feature extraction, which guides the model to focus on the task-related features of the foreground objects, and suppress much noise due to the lack of supervised knowledge. Moreover, we introduce multiple meta-prototypes to encode hierarchical features and learn class-agnostic structural information. The hierarchical features help the model highlight the decision boundary and focus on hard pixels, and the structural information learned from base classes is treated as the prior knowledge for novel classes. Experiments show that our method achieves the mean-IoU scores of 60.79\% and 41.16\% on PASCAL-$5^i$ and COCO-$20^i$, outperforming the state-of-the-art method by 3.49\% and 5.64\% in the 5-shot setting. Moreover, comparing with 1-shot results, our method promotes 5-shot accuracy by 3.73\% and 10.32\% on the above two benchmarks. The source code of our method is available at \url{https://github.com/Jarvis73/PEMP}.
\end{abstract}

\section{Introduction}
Deep convolutional neural networks have extensively promoted semantic segmentation accuracy in past years~\cite{longFullyConvolutionalNetworks2015, ronnebergerUnetConvolutionalNetworks2015, chenDeepLabSemanticImage2018, chenEncoderDecoderAtrousSeparable2018, zhaoPyramidSceneParsing2017}. The inspiring performances usually profit from a supervised manner at the price of huge amounts of densely annotated images~\cite{everinghamPascalVisualObject2015, linMicrosoftCOCOCommon2014}. However, collecting pixel-wise object labels at scale is notoriously known as cumbersome, expensive, and sometimes impractical when developing models for novel scenarios. To tackle this challenge, few-shot segmentation~(FSS)~\cite{shaban2017one, rakelly2018conditional, Siam_2019_ICCV} is developed for image segmentation with a few labeled samples and can generalize to unseen classes quickly. 

FSS is derived from the few-shot learning (FSL)~\cite{vinyalsMatchingNetworksOne2016, snellPrototypicalNetworksFewshot2017, sungLearningCompareRelation2018}. Unlike the standard supervised training scheme that models are trained and evaluated on the same classes, FSL models are trained on base classes with fully labeled images and evaluated on novel classes with only a few labeled samples. Recent FSS methods~\cite{Zhang_2019_CVPR, Wang_2019_ICCV, zhangSGOneSimilarityGuidance2020, gairolaSimPropNetImprovedSimilarity2020, yangPrototypeMixtureModels2020} generally follow an episodic meta-training scheme where the training and testing samples are formulated as tasks, \ie, \textit{episodes}. Precisely, each episode consists of a few labeled {\it support} images and a {\it query} image. Support images and labels are adopted to extract task-related features, also known as the ``prototypes'', which are then leveraged to segment the query image.

Nevertheless, despite the achieved improvements, the meta-learning based approaches particularly suffer from three limitations: 
{\bf(1) Models are distracted by task-unrelated information.} 
In the inference stage of FSS, as the support labels are utilized {\it after} the feature extraction, models have no knowledge about the novel class when extracting features. Therefore, the features contain much noise, which distracts models from task-related information.
{\bf(2) The representation ability of a single prototype is limited.} 
Using a single prototype to represent a class is widely adopted in recent work~\cite{zhangSGOneSimilarityGuidance2020, Wang_2019_ICCV, tianPriorGuidedFeature2020}. However, the prototype is obtained by global average pooling, which deteriorates the feature distribution~\cite{yangPrototypeMixtureModels2020} and degenerates the decision boundary. 
{\bf(3) Class-related prototypes ignore the prior knowledge.} 
In the inference stage, previous methods~\cite{Zhang_2019_CVPR, Wang_2019_ICCV, yangPrototypeMixtureModels2020} generate prototypes only utilizing the support features of the novel class. Therefore, the class-related prototypes do not benefit from the prior knowledge of the base classes. 

This paper proposes the Prior-Enhanced network with multiple Meta-Prototypes~(PEMP), which contains two parts to overcome the above limitations. The prior-enhanced network is designed to handle the first limitation by leveraging support labels in support feature extraction, which guides the model to focus on foreground objects and reduces task-unrelated information. We introduce a prior network to generate query pseudo-labels, which allows the query features also benefit from the prior-enhanced network. Moreover, since the support and query features are extracted in two separated branches, a {\it Communication Module} is designed to establish communication between the two branches for selecting the most discriminate features~\cite{Hao_2019_ICCV} of the foreground objects.
To tackle the remaining two limitations, we introduce multiple {\it meta-prototypes} to encode hierarchical features for highlighting the decision boundary and learn structural information for aggregating basic knowledge. Concretely, features in the embedding space are represented by multiple prototypes according to the relative distances between the features and the decision boundary. Therefore, models are encouraged to pay more attention to the features close to the decision boundary, improving the accuracy of hard pixels. We extend standard class-related prototypes by class-agnostic meta-prototypes, which aggregate knowledge of base classes by learning structural information from the training set. The structural information corresponds to the relative distances between features and the decision boundary existing in all the base classes.

We evaluate the proposed method on two benchmarks PASCAL-$5^i$~\cite{shaban2017one} and COCO-$20^i$~\cite{linMicrosoftCOCOCommon2014}. The results demonstrate that our method achieves state-of-the-art performance in the 1-shot setting and outperforms state-of-the-arts with a large margin in the 5-shot setting. To sum up, the main contributions of the work are summarized as follows:
\begin{itemize}
    \item We develop PEMP for few-shot segmentation. We propose a prior-enhanced network to leverage support and query (pseudo-) labels in feature extraction, which guides the model to focus on task-related features and suppress noise information.
    \item We propose a communication module to break the isolation between the support branch and query branch, which help select the most discriminative features.
    \item We introduce multiple meta-prototypes that encode hierarchical features to highlight the decision boundary and aggregate prior knowledge of base classes.
    \item Experiments on PASCAL-$5^i$ and COCO-$20^i$ show that our method significantly outperforms the state-of-the-art performance in the 5-shot setting and can effectively improve segmentation accuracy with more support images. 
\end{itemize}

\section{Related Work}

\paragraph{Semantic segmentation.} Semantic segmentation problem has been extensively investigated for decades. State-of-the-art methods extend the fully convolutional networks~(FCN)~\cite{longFullyConvolutionalNetworks2015} by leveraging skip connections~\cite{ronnebergerUnetConvolutionalNetworks2015}, atrous convolution~\cite{chenEncoderDecoderAtrousSeparable2018} or pyramid pooling~\cite{zhaoPyramidSceneParsing2017}, which are widely adopted in segmentation problems. Based on the atrous convolution, Chen \etal~\cite{chenDeepLabSemanticImage2018} propose an Atrous Spatial Pyramid Pooling (ASPP) layer~\cite{chenEncoderDecoderAtrousSeparable2018}, which is applied in semantic segmentation~\cite{chenDeepLabSemanticImage2018, chenEncoderDecoderAtrousSeparable2018} and FSS~\cite{Zhang_2019_CVPR, yangPrototypeMixtureModels2020}. In this work, we follow the structure of FCN with atrous convolution to learn features in the few-shot semantic segmentation. With the extracted features, the prediction is generated based on a metric function as in PANet~\cite{Wang_2019_ICCV}, which is different from the scheme of standard semantic segmentation.

\paragraph{Few-shot segmentation.}

Shaban \etal~\cite{shaban2017one} introduce few-shot learning into semantic segmentation tasks with a dense prediction strategy, formulating the few-shot segmentation scheme. To sufficiently utilize the supervised knowledge, Zhang \etal~\cite{zhangSGOneSimilarityGuidance2020} propose to generate prototypes by Masked Average Pooling on support features with corresponding labels. Besides, the places of support and query features can be exchanged in feature space because they belong to the same class. Therefore, Wang \etal~\cite{Wang_2019_ICCV} implement prototype alignment between support and query features by adding an extra loss on support predictions. Zhang \etal~\cite{Zhang_2019_CVPR} propose to refine segmentation results by an iterative optimization module, which can recurrently fine-tune the segmentation results. Yang \etal~\cite{yangPrototypeMixtureModels2020} insert a prototype mixture model on top of the feature extractor. The mixture model generates multiple prototypes representing semantic parts of the foreground objects. Nevertheless, the concrete meaning of each prototype is too ambiguous to explain. 

\paragraph{Prior knowledge.}

\begin{figure*}[t]
\begin{center}
  \includegraphics[width=\linewidth]{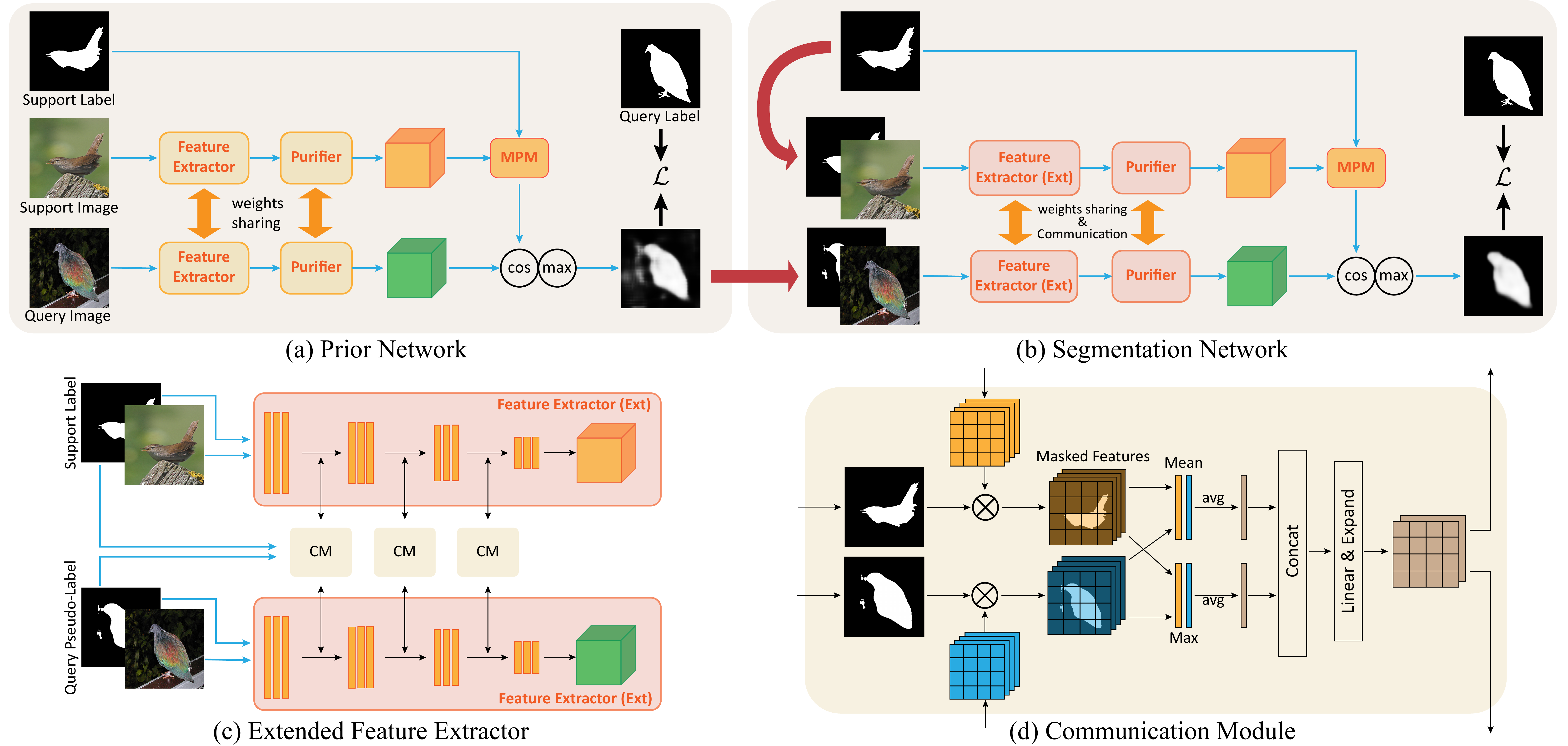}
\end{center}
  \caption{Overview of the proposed two-stage Prior-Enhanced network with Meta-Prototypes. (a) The Prior network~(the first stage) is employed to provide query pseudo-labels. (b) The Segmentation network~(the second stage) is responsible for making the final prediction by leveraging (pseudo-) labels in feature extraction. (c) Details of the extended feature extractor with (pseudo-) labels. (d) The Communication Module for fusing support features and query features. MPM: Meta-Prototype Module; CM: Communication Module.}
\label{fig:framework}
\end{figure*}

Prior knowledge has been widely used in semantic segmentation. Yu \etal~\cite{yuContextPriorScene2020} propose Context Prior to encode the relationships between pixels. The Context Prior Map is supervised with an Ideal Affinity Map computed from segmentation labels. Cheng \etal~\cite{chengCascadePSPClassAgnosticVery2020} address the high-resolution segmentation problem with cascade refinement modules. The input of each module is the prior knowledge from the previous module. Prior knowledge is also investigated in the FSS problem. Tian \etal~\cite{tianPriorGuidedFeature2020} propose to generate a prior map from high-level features based on a pre-trained feature extractor. Then, the prior map is concatenated with middle-level features and fed into a feature enhancement module. Pambala \etal~\cite{pambalaSMLSemanticMetalearning2020} introduce a semantic meta-learning method that incorporates class-level semantic description into the prototypes. However, the semantic learning scheme requires an extra embedding dataset that is not always available in practice. Some efforts have been made to utilize the prior knowledge of the base classes~\cite{Siam_2019_ICCV, gidarisDynamicFewShotVisual2018, liuDynamicExtensionNets2020}, where a classifier encodes the class-specific knowledge. However, the classifier size linearly increases with the number of base classes, which results in unlimited model size. Therefore, we propose the meta-prototypes to summarize the knowledge of base classes into a fixed number of feature vectors, which efficiently controls the classifier size.

\section{Method}

\textbf{Problem Setting } Following the training scheme of meta-learning~\cite{vanschorenMetaLearningSurvey2018}, FSS methods perform training with a series of episodes. Each episode is formed as an $N$-class segmentation task with $K$ support (image, label) pairs ($x^s$, $y^s$) in each class and a query (image, label) pair ($x^q$, $y^q$). The problem is formulated as an $N$-way $K$-shot FSS problem where $K$ is a small number. In this work, we focus on the $1$-way $K$-shot problem.

\subsection{Baseline}\label{sec:baseline}
We first set a Baseline model, which is a vanilla model of the metric-based FSS methods.
Given a support image $x^s$ and a query image $x^q$, we embed them into a feature space $\mathcal{F}$ with a feature extractor $f_{\theta}$, where $\theta$ denotes trainable parameters and $f_{\theta}(x^s), f_{\theta}(x^q)\in\mathcal{F}\subset\mathbb{R}^{c\times h\times w}$, where $c$, $h$, $w$ denote the channel, height and width of the features, respectively. Let $h^s_i,h^q_i\in\mathbb{R}^c$ be the feature vectors of $f_{\theta}(x^s)$ and $f_{\theta}(x^q)$, respectively, at the spatial location $i\in\{1,2,\cdots,HW\}$. With the support labels $y^s_i\in\{0, 1\}$, we have foreground and background prototypes as follows:
\begin{equation*}
    p_{FG} = \frac{\sum_i h^s_i\cdot y^s_i}{\sum_i y^s_i},\;\; p_{BG} = \frac{\sum_i h^s_i \cdot (1 - y^s_i)}{HW - \sum_i y^s_i}.
\end{equation*}
The query prediction is obtained by computing the cosine similarity between the $h^q_i$ and $p_r$, where $r\in \{FG, BG\}$. Then we get prediction maps by
\begin{equation*}
    \hat{y}_i = \underset{r}{\mbox{softmax}}(\gamma d_s(h^q_i, p_r)),
\end{equation*}
where $d_s(\cdot,\cdot)$ denotes the cosine similarity function, and $\gamma$ is a hyper-parameter controlling the distance scale.

\subsection{Prior-Enhanced Network}\label{sec:structure}
We present the prior-enhanced network to boost the task-related features by capitalizing on support and query (pseudo-) labels in feature extraction. As shown in Fig.~\ref{fig:framework}, the prior-enhanced network is a two-stage model, consisting of a prior network and a segmentation network. Moreover, we design a {\it Communication Module} to exchange information between the support and query branches, which further enhances the foreground object features. 

\paragraph{Two-stage networks} Since query labels are unavailable in the inference stage, we propose the prior network to generate prior maps, which can be transformed into pseudo-labels, as shown in Fig.~\ref{fig:framework}~(a). We first embed support and query images into a low-dimension feature space according to a weight-shared feature extractor and a purifier. Then, we learn multiple prototypes from support features by the {\it Meta-Prototype Module} (MPM) (\textit{cf.}~\S\ref{sec:MPM}). The cosine similarity function measures the distances of the prototypes and query features, and the resulting distance maps are merged as a query prior map. Finally, we binarize the prior map as a pseudo-label that is consequently fed into the segmentation network.

From the prior network, we get the pseudo-labels for query images. The segmentation network takes both images and (pseudo-) labels as input for paying attention on task-related objects, as shown in Fig.~\ref{fig:framework}~(b). We extend the feature extractor by inserting extra convolutional kernels into multiple layers to leverage (pseudo-) labels, as shown in Fig.~\ref{fig:framework}~(c). The (pseudo-) labels are incorporated into the input layer as an extra channel, which is a strong prior for extracting crucial features. They are also used in {\it Communication Modules} to extract the most discriminative foreground features.

When the feature extractor~(\eg, Resnet~\cite{heDeepResidualLearning2016}) includes residual paths, the generated features contain too much noise because the residual paths bring low-level features into high layers. Therefore, we introduce a purifier on top of the feature extractor to alleviate this problem. The purifier consists of two convolutional layers and an ASPP layer~\cite{chenEncoderDecoderAtrousSeparable2018} for reducing noise and enlarging the receptive field of the output features. 

\paragraph{Communication Modules} The idea of feature communication is inspired by the work of low-light enhancement where the features of multi-exposure images are fused~\cite{zhuEEMEFNLowLightImage2020}. {\it Communication Modules} are applied before each convolutional block for guiding the model to extract the most discriminative features. As shown in Fig.~\ref{fig:framework}~(d), given the intermediate support and query features $\tilde{h}^s_i, \tilde{h}^q_i$, and the downsampled (pseudo-) labels $\tilde{y}^s, \tilde{y}^q$, we take the mean and max features on the spatial dimensions within object regions:
\begin{equation*}
    o_{\mbox{\small mean}}^t =\frac1{HW}\sum_i\tilde{h}^t_i\cdot \tilde{y}^t_i,\quad o_{\mbox{\small max}}^t =\max_i\tilde{h}^t_i\cdot \tilde{y}^t_i,
\end{equation*}
where $t\in\{s, q\}$. We perform a ``merge-and-distribute'' strategy to achieve communication between features. Support and query features are first merged and processed by
\begin{equation*}
    u = W\left(\left[\frac{o^s_{\mbox{\small mean}}+o^q_{\mbox{\small mean}}}{2}\;\; \frac{o^s_{\mbox{\small max}}+o^q_{\mbox{\small max}}}{2}\right]\right) + b,
\end{equation*}
where $W$ and $b$ are weights and biases of a linear layer, and $[\;]$ is the concatenate operation. The linear layer compresses hundreds of features to only two features, which avoids affecting the original features too much. Then, we copy $u$ back to the support and query branches to finish the distribution step. 

\subsection{Meta-Prototypes Module}\label{sec:MPM}

\begin{figure}[t]
\begin{center}
  \includegraphics[width=\linewidth]{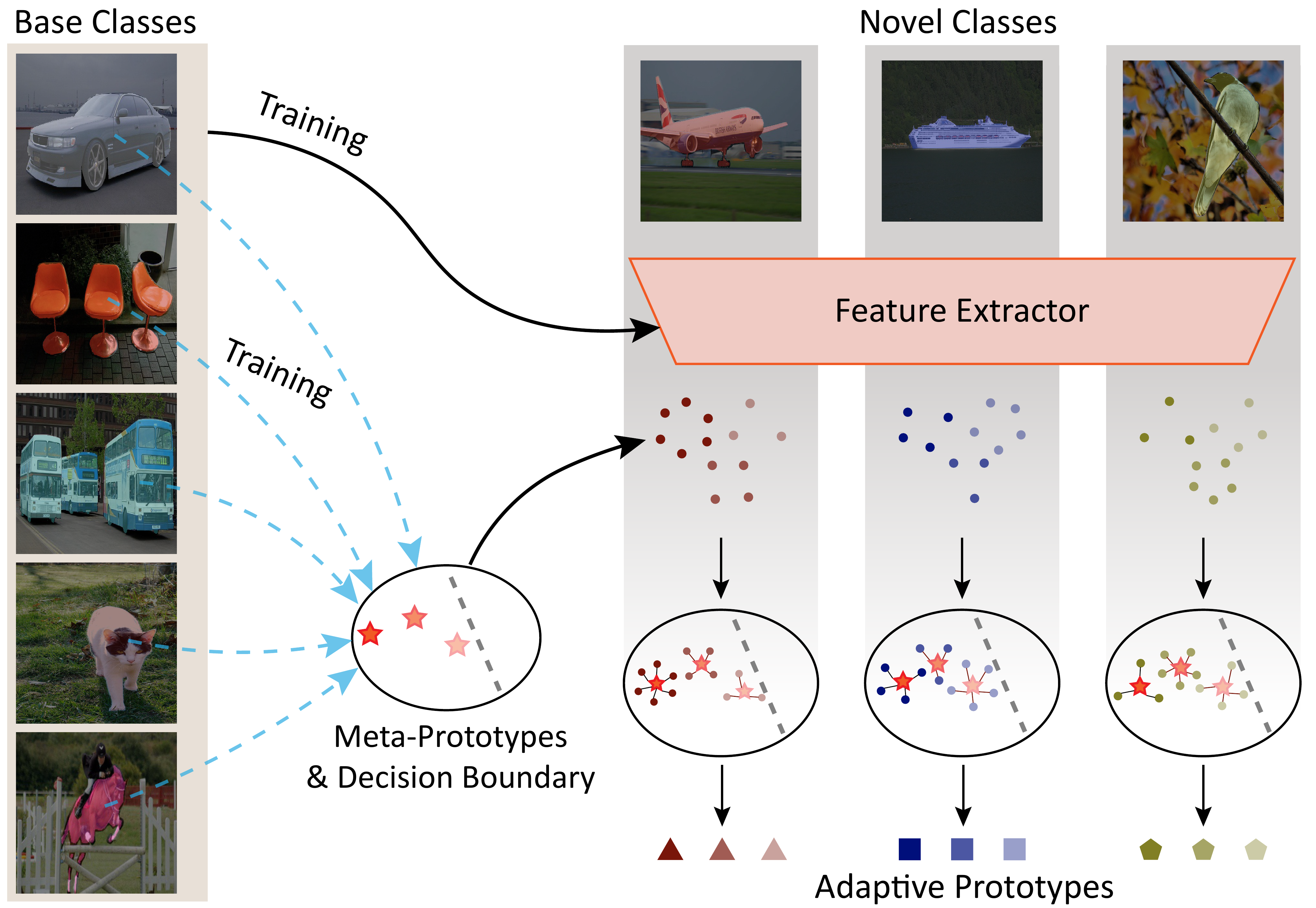}
\end{center}
  \caption{Illustration of the proposed multiple meta-prototypes. Meta-prototypes summarize prior knowledge from all the base classes and then are combined with features of novel classes to generate adaptive prototypes. Multiple prototypes are used to encode hierarchical features to highlight the decision boundary.}
\label{fig:header}
\end{figure}

We introduce multiple meta-prototypes to better identifying the decision boundary and aggregate prior knowledge from the base classes. As shown in Fig.~\ref{fig:header}, the knowledge of training images is learned in the feature extractor while the prior knowledge of base classes is aggregated into multiple meta-prototypes. With a specific task, meta-prototypes are transformed into {\it adaptive prototypes} and the hierarchical features highlight the decision boundary. 

Let $q_m^r$ denote meta-prototypes with $r\in\{FG, BG\}$ and $m=1,2,\cdots,M$, where $M$ is the number of prototypes for each class. We first measure the similarity between support features $h_i^s$ and each meta-prototype by the Euclidean distance $d_e(h_i^s, p_m^r)$. Then, we apply softmax to get attention coefficients of the support features with each meta-prototype as soft assignments:
\begin{equation*}
    \alpha_{i,m}^r = \frac{\exp\left(d_e(h_i^s, q_m^r)\right)}{\sum_{l=1}^M\exp\left(d_e(h_i^s, q_l^r)\right)},
\end{equation*}
which push feature vectors to the closest meta-prototype. The support features are scaled with the coefficients and averaged over the class-specific spatial regions, yielding adaptive prototypes:
\begin{equation*}
    \hat{p}^r_m = \frac1{\vert\mathcal{I}^r\vert}\sum_{i\in\mathcal{I}^r}x_i^s\cdot\alpha_{i,m}^r,
\end{equation*}
where $\mathcal{I}^r$ denotes the index set of either the objects or the backgrounds, and $\vert\cdot\vert$ gives the number of elements. 

Finally, multiple query predictions are obtained by measuring the similarities between the query features and multiple adaptive prototypes. We utilize a {\it max} operation to merge the predictions:
\begin{equation*}
    \hat{y}_i = \underset{r}{\mbox{softmax}}(\gamma \max_m d_s(h^q_i, \hat{p}^r_m)),
\end{equation*}
which is either the query prior map for the prior network or the query prediction for the segmentation network.

\subsection{Training Objective} 

The proposed PEMP is optimized with the cross-entropy loss in two stages. We first train the prior network to predict query prior maps. Then, we fix the prior network parameters, and train the segmentation network for final segmentation results. 
We introduce a boundary-enhanced weight map for the cross-entropy loss. We assign large weights to the pixels close to object boundaries and small weights to other pixels with an exponential decay function, . Specifically, let $B$ denote a binary map where only object boundary pixels are set to the positive value. The weight map is formulated as
\begin{equation}\label{eq:weight}
    w = \exp\left(\frac{-\mbox{EDT}(\neg B)}{\sigma^2}\right) + 1,
\end{equation}
where $\mbox{EDT}$ is the Euclidean distance transform~\cite{xuEuclideanDistanceTransform2006}, and $\neg$ is the boolean {\it not} operation. The $\sigma$ is a hyper-parameter that adjusts the decay rate of weights from boundary pixels to other pixels. The constant one is added to avoid zero weights for the pixels far from the boundaries. The final loss function is
\begin{equation*}
    \mathcal{L} = \frac1{HW}\sum_iw_i(y_i\log\hat{y}_i + (1-y_i)\log(1-\hat{y}_i)),
\end{equation*}
which is used to train PEMP.

\section{Experiments}

\subsection{Setup}
\paragraph{Datasets} 

We conduct experiments on two datasets: PASCAL-$5^i$~\cite{shaban2017one} and COCO-$20^i$~\cite{Nguyen_2019_ICCV}. PASCAL-$5^i$ is built from the PASCAL VOC 2012~\cite{pascal-voc-2012} and extended annotations from BSDS~\cite{hariharanSemanticContoursInverse2011}. It consists of more than 15,000 annotated images of 20 classes. The dataset is divided into four splits (5 classes per split) following the label order for cross-validation. 
COCO-$20^i$ is built from MS COCO 2014 dataset~\cite{linMicrosoftCOCOCommon2014}. It contains more than 80,000 images for training and more than 40,000 images for evaluation of 80 classes. Similar to PASCAL-$5^i$, we divide the set of classes into four splits. For the details of the classes in each split, please refer to the work of Nguyen \etal~\cite{Nguyen_2019_ICCV}. 

\paragraph{Evaluation metrics}
We use the mean-IoU and binary-IoU~\cite{Wang_2019_ICCV} to evaluate our method. The mean-IoU computes average intersection-over-union~(IoU) over all the novel classes, while the binary-IoU combines all the novel classes into a single foreground class and takes the average value between the foreground class and the background class. Following PANet~\cite{Wang_2019_ICCV}, we evaluate the model with 1000 random episodes at each testing run. The average score of 5 runs are reported for a stable result.

\begin{table*}[!t]
\centering
\begin{tabular}{l|c|ccccc|ccccc|c}
\toprule
\multirow{2}{*}{Method} & \multirow{2}{*}{FE} & \multicolumn{5}{c|}{1-shot} & \multicolumn{5}{c|}{5-shot} & \multirow{2}{*}{$\Delta$} \\ \cline{3-12}
& & split-0 & split-1 & split-2 & split-3 & Mean & split-0 & split-1 & split-2 & split-3 & Mean & \\
\midrule
OSLSM\cite{shaban2017one} & \multirow{8}{*}{\rotatebox{90}{VGG-16}} & 33.60 & 55.30 & 40.90 & 33.50 & 40.80 & 35.90 & 58.10 & 42.70 & 39.10 & 43.95 & 3.15 \\
co-FCN~\cite{rakelly2018conditional} & & 36.70 & 50.60 & 44.90 & 32.40 & 41.10 & 37.50 & 50.00 & 44.10 & 33.90 & 41.40 & 0.30 \\
SG-One~\cite{zhangSGOneSimilarityGuidance2020} & & 40.20 & 58.40 & 48.40 & 38.40 & 46.30 & 41.90 & 58.60 & 48.60 & 39.40 & 47.10 & 0.80 \\
PANet~\cite{Wang_2019_ICCV} & & 42.30 & 58.00 & 51.10 & 41.20 & 48.10 & 51.80 & 64.60 & 59.80 & 46.50 & 55.70 & 7.60 \\
FWB~\cite{Nguyen_2019_ICCV} & & 47.04 & 59.64 & 52.61 & 48.27 & 51.90 & 50.87 & 62.86 & 56.48 & 50.09 & 55.08 & 3.18 \\
RPMMs~\cite{yangPrototypeMixtureModels2020} & & 47.14 & {\bf 65.82} & 50.57 & {\bf 48.54} & 53.02 & 50.00 & 66.46 & 51.94 & 47.64 & 54.01 & 0.99 \\ 
Baseline & & 42.69 & 56.43 & 47.59 & 42.19 & 47.23 & 51.91 & 63.98 & 58.84 & 48.40 & 55.78 & {\bf 8.55} \\
PEMP (Ours) & & {\bf 50.05} & 64.22 & {\bf 53.65} & 45.97 & {\bf 53.47} & {\bf 57.10} & {\bf 68.46} & {\bf 62.54} & {\bf 55.06} & {\bf 60.79} & 7.32 \\ \hline
CaNet$^{\dagger}$~\cite{Zhang_2019_CVPR} & \multirow{5}{*}{\rotatebox{90}{Resnet-50}} & 53.21 & 64.07 & 49.58 & 50.98 & 54.46 & 55.12 & 66.57 & 50.42 & 53.00 & 56.28 & 1.82 \\
PFENet$^{\dagger}$~\cite{tianPriorGuidedFeature2020} & & {\bf 55.89} & 65.89 & 49.03 & {\bf 53.10} & 55.98 & 58.00 & 67.21 & 49.31 & 52.73 & 56.81 & 0.83 \\
RPMMs$^{\dagger}$~\cite{yangPrototypeMixtureModels2020} & & 53.86 & {\bf 66.45} & 52.76 & 51.31 & 56.10 & 56.28 & 67.34 & 54.52 & 51.00 & 57.30 & 1.20 \\
Baseline & & 45.48 & 59.97 & 51.35 & 43.31 & 50.03 & 52.47 & 66.31 & 59.85 & 51.02 & 57.41 & {\bf 7.38} \\
PEMP (Ours) & & 55.74 & 65.88 & {\bf 54.12} & 50.34 & {\bf 56.52} & {\bf 58.59} & {\bf 69.10} & {\bf 60.31} & {\bf 53.01} & {\bf 60.25} & 3.73 \\
\bottomrule
\end{tabular}
\caption{Mean-IoU results of 1-shot and 5-shot segmentation on the PASCAL-$5^i$. The results marked by $\dagger$ are re-evaluated on the same validation pairs as ours. $\Delta$ denotes the differences between the 1-shot and 5-shot results. FE: Feature extractor.} \label{tab:pascal-mean}
\end{table*}

\begin{table}[!t]
\centering
\begin{tabular}{l|c|cc|c}
\toprule
Method & FE & 1-shot & 5-shot & $\Delta$ \\
\midrule
OSLSM\cite{shaban2017one} & \multirow{6}{*}{VGG-16} & 61.30 & 61.50 & 0.20 \\
co-FCN~\cite{rakelly2018conditional} & & 60.10 & 60.20 & 0.10 \\
SG-One~\cite{zhangSGOneSimilarityGuidance2020} & & 63.90 & 65.90 & 2.00 \\
PANet~\cite{Wang_2019_ICCV} & & 66.50 & 70.70 & 4.20 \\
Baseline & & 65.51 & 70.24 & 4.73 \\
PEMP (Ours) & & {\bf 70.50} & {\bf 75.69} & {\bf 5.19} \\ \hline
CaNet$^{\dagger}$~\cite{Zhang_2019_CVPR} & \multirow{5}{*}{Resnet-50} & 67.23 & 69.58 & 2.35 \\
PFENet$^{\dagger}$~\cite{tianPriorGuidedFeature2020} & & {\bf 71.42} & 72.63 & 1.21 \\
RPMMs$^{\dagger}$~\cite{yangPrototypeMixtureModels2020} & & 70.32 & - & - \\
Baseline & & 67.58 & 71.90 & {\bf 4.32} \\
PEMP (Ours) & & 71.41 & {\bf 73.84} & 2.43 \\
\bottomrule
\end{tabular}
\caption{Binary-IoU results of 1-shot and 5-shot segmentation on the PASCAL-$5^i$. The results marked by $\dagger$ are re-evaluated on the same validation pairs as ours. $\Delta$ denotes the differences between the 1-shot and 5-shot results. FE: Feature extractor.} \label{tab:pascal-binary}
\end{table}

\paragraph{Implementation details}

We evaluate our method with VGG-16~\cite{simonyanVeryDeepConvolutional2014} and Resnet-50~\cite{heDeepResidualLearning2016} as the feature extractor. We omit the purifier when using the VGG-16 feature extractor as no residual paths exist. The weights of the feature extractor are pre-trained on ImageNet~\cite{dengImageNetLargescaleHierarchical2009} and fine-tuned on the training set as in previous work~\cite{Wang_2019_ICCV,yangPrototypeMixtureModels2020}. Input images are resized to $401\times 401$ and augmented with random flipping, random resizing, and color jitter. SGD is used to train the model with the momentum of 0.9. To accelerate and stabilize the training process, we apply gradient clipping with a norm of 1.1 before updating parameters in each training step~\cite{zhangWhyGradientClipping2019}. We train the prior network for 90 epochs with a fixed learning rate of 0.001~\cite{Wang_2019_ICCV} and the segmentation network for 200 epochs with a fixed learning rate of 0.0035~\cite{yangPrototypeMixtureModels2020}. The weight decay is 0.0005, and the batch size is 4. We set the hyper-parameter $\gamma$ to 20 as in PANet~\cite{Wang_2019_ICCV}. The number of prototypes $M$ is 3, and the scale factor $\sigma$ is 5.0 by default.

\paragraph{Baseline}

We implement a Baseline model for highlighting the contribution of the proposed method. The Baseline model is a vanilla FSS method without the prior-enhanced strategy, the purifier, and multiple meta-prototypes. Details of the Baseline is described in Sec.~\ref{sec:baseline}. 

\subsection{Results and Discussion}\label{sec:results}
\subsubsection{PASCAL-$5^i$} 

\paragraph{Quantitative results}
Table~\ref{tab:pascal-mean} displays the comparison among PEMP and previous works. In the 1-shot setting, compared with the Baseline, PEMP achieves significant improvements by 6.24\% and 6.49\% with VGG-16 and Resnet-50 feature extractor, respectively. Compared with state-of-the-arts, PEMP achieves modest improvements. In the 5-shot setting, PEMP~(VGG-16) significantly outperforms the Baseline by 5.01\% and outperforms the state-of-the-art method RPMMs~(Resnet-50) by 3.49\%. Meanwhile, PEMP consistently improves the accuracy in all the splits. The gains from 1-shot to 5-shot are limited for most previous methods. But PEMP gains by 7.32\% and 3.73\% with the above two feature extractors, respectively. It benefits from the guidance of the supervised knowledge and the meta-prototypes that allows the integration of basic knowledge.
Table~\ref{tab:pascal-binary} display the results in binary-IoU. PEMP outperforms state-of-the-arts in most of the settings. For the 1-shot setting with Resnet-50, PEMP still achieves a comparable result with PFENet~\cite{tianPriorGuidedFeature2020}.

PEMP makes prediction with a non-parametric cosine similarity function as in PANet~\cite{Wang_2019_ICCV}. Previous methods achieving the top performance, such as CaNet~\cite{Zhang_2019_CVPR} and RPMMs~\cite{yangPrototypeMixtureModels2020}, make prediction by relation networks. In this view, the experiments indicate that non-parametric predictors can also achieve state-of-the-art performance, suggesting that learning a good embedding is important for FSS. 

\begin{figure*}[t]
\begin{center}
  \includegraphics[width=\linewidth]{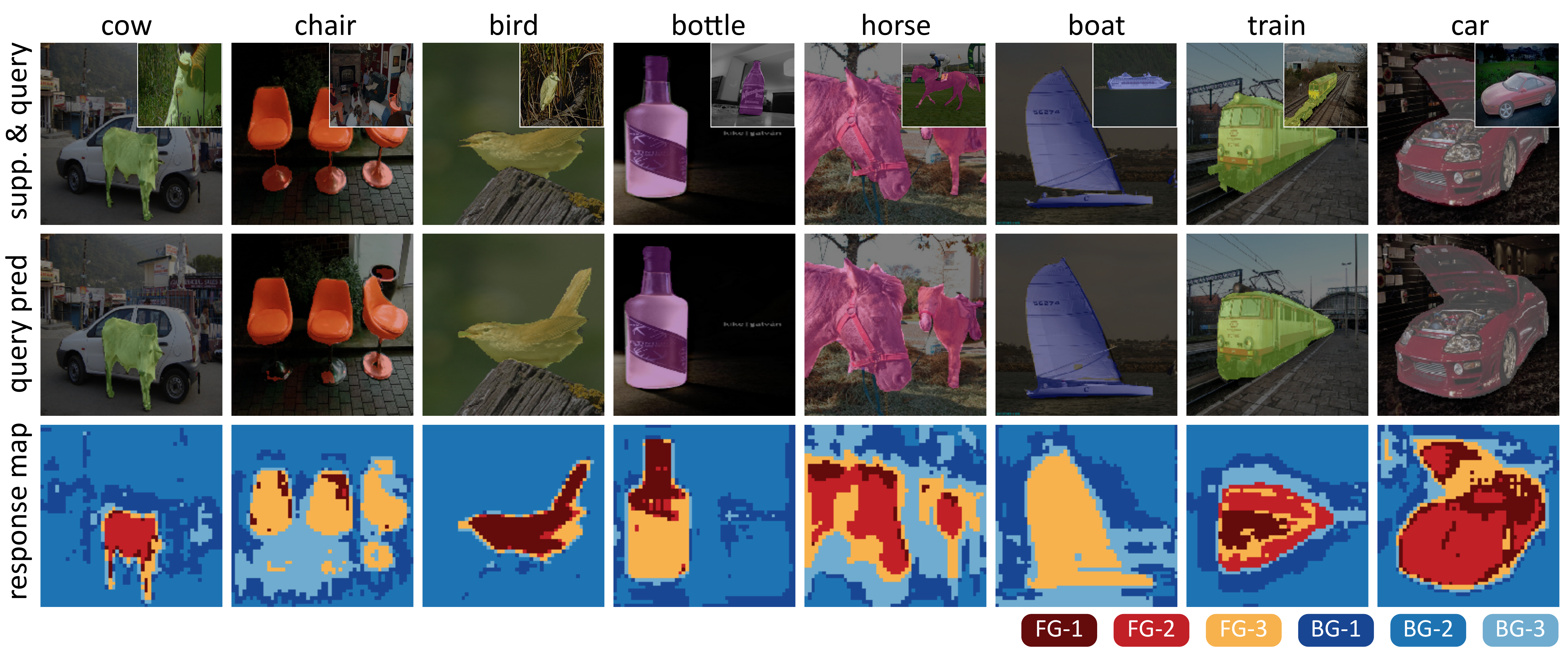}
\end{center}
  \caption{Samples of the segmentation results with the proposed method on the Pascal-$5^i$ dataset and 1-shot setting. The first row shows the support images~(at the corner) and query images. The second row shows the query predictions, and the last row presents response maps of the query features to multiple prototypes. The meaning of the colors in response maps is denoted in the right bottom corner.}
\label{fig:pascal}
\end{figure*}

\paragraph{Qualitative results}
We present some qualitative results of our method in Fig.~\ref{fig:pascal}. The first row gives support and query images from different classes, and the second row is the predictions, which indicates PEMP can produce accurate segmentation results even with one support image. The last row presents the response maps to multiple foreground prototypes and background prototypes. They are generated from the indices of the {\it max} operation.
From the response maps, one can observe that an object can be captured by one prototype~(\eg, the boat) or multiple prototypes~(\eg, the bird), which depends on the similarity of the support and query objects. The response maps reflect the hierarchical features encoded by multiple meta-prototypes. Specifically, for query pixels determinedly belonging to the object~(\eg, inner pixels of the bird), they have a higher response to foreground prototypes FG-1 and FG-2. For those pixels less similar to the object~(\eg, border pixels of the bird and all the boat pixels), they are close to the decision boundary and have a higher response to prototype FG-3. Background pixels also suggest the hierarchical structure~(\eg, the train). With multiple meta-prototypes, PEMP can recognize {\it easy} pixels with high confidence and pay more attention to {\it hard} pixels close to the decision boundary. 

\begin{table*}[!t]
\centering
\begin{tabular}{c|l|ccccc|ccccc|c}
\toprule
\multirow{2}{*}{Met.} & \multirow{2}{*}{Method} & \multicolumn{5}{c|}{1-shot} & \multicolumn{5}{c|}{5-shot} & \multirow{2}{*}{$\Delta$} \\ \cline{3-12}
& & split-0 & split-1 & split-2 & split-3 & Mean & split-0 & split-1 & split-2 & split-3 & Mean \\
\midrule
\multirow{3}{*}{\rotatebox{90}{m-IoU}} & PANet~\cite{Wang_2019_ICCV} & - & - & - & - & 20.90 & - & - & - & - & 29.70 & 8.80 \\
& FWB~\cite{Nguyen_2019_ICCV} & 16.98 & 17.98 & 20.96 & 28.85 & 21.19 & 19.13 & 21.46 & 23.93 & 30.08 & 23.65 & 2.46 \\
& RPMMs~\cite{yangPrototypeMixtureModels2020} & {\bf 29.53} & {\bf 36.82} & 28.94 & 27.02 & 30.58 & 33.82 & 41.96 & 32.99 & 33.33 & 35.52 & 4.94 \\
& PEMP (Ours) & 29.28 & 34.09 & {\bf 29.64} & {\bf 30.36} & {\bf 30.84} & {\bf 39.08} & {\bf 44.59} & {\bf 39.54} & {\bf 41.42} & {\bf 41.16} & {\bf 10.32} \\ \hline
\multirow{3}{*}{\rotatebox{90}{b-IoU}} & A-MCG~\cite{huAttentionBasedMultiContextGuiding2019} & - & - & - & - & 52.00 & - & - & - & - & 54.70 & 2.70 \\
& PANet~\cite{Wang_2019_ICCV} & - & - & - & - & 59.20 & - & - & - & - & 63.50 & 4.30 \\
& PEMP (Ours) & {\bf 60.41} & {\bf 64.46} & {\bf 64.49} & {\bf 63.14} & {\bf 63.13} & {\bf 67.73} & {\bf 72.48} & {\bf 70.05} & {\bf 72.56} & {\bf 70.71} & {\bf 7.58} \\
\bottomrule
\end{tabular}
\caption{Comparison of the proposed method and state-of-the-art methods on the MS COCO-$20^i$ dataset. All the results are acquired based on the Resnet-50 feature extractor. $\Delta$ denotes the difference between the 5-shot results and the corresponding 1-shot results. Met.: Metric.} \label{tab:coco}
\end{table*}

\subsubsection{MS COCO-$20^i$}

The results of the COCO-$20^i$ dataset are shown in Table~\ref{tab:coco}. COCO-$20^i$ has more fine-grained categories than PASCAL-$5^i$, and is more challenging. For the results in mean-IoU, PEMP marginally outperforms state-of-the-arts in the 1-shot setting and makes a significant improvement of 5.64\% in the 5-shot setting. We list the results in binary-IoU for reference. Like the observation in PASCAL-$5^i$ experiments, PEMP achieves a remarkable improvement by 10.32\% when more support images are available. 

\subsection{Ablation Study}\label{sec:ablation}

\paragraph{Structure analysis} We conduct ablation experiments to display the contribution of each module as shown in Table~\ref{tab:structure}. The output of the prior network is not only a pseudo-label but a query prediction. With the purifier, the prior network achieves an improvement of 2.03\% comparing with the Baseline. When introducing the segmentation network, the accuracy is further promoted by 1.69\%. Meta-prototypes achieve apparent improvements by 1.77\% and 2.34\% in the prior network and the segmentation network, respectively. Moreover, with {\it Communication Modules}, the accuracy is further promoted by 0.43\% and achieves 56.52\%. These experiments suggest that the proposed modules consistently improve prediction accuracy, and their contributions are effectively accumulated. 

\begin{figure}[t]
\begin{center}
  \includegraphics[width=\linewidth]{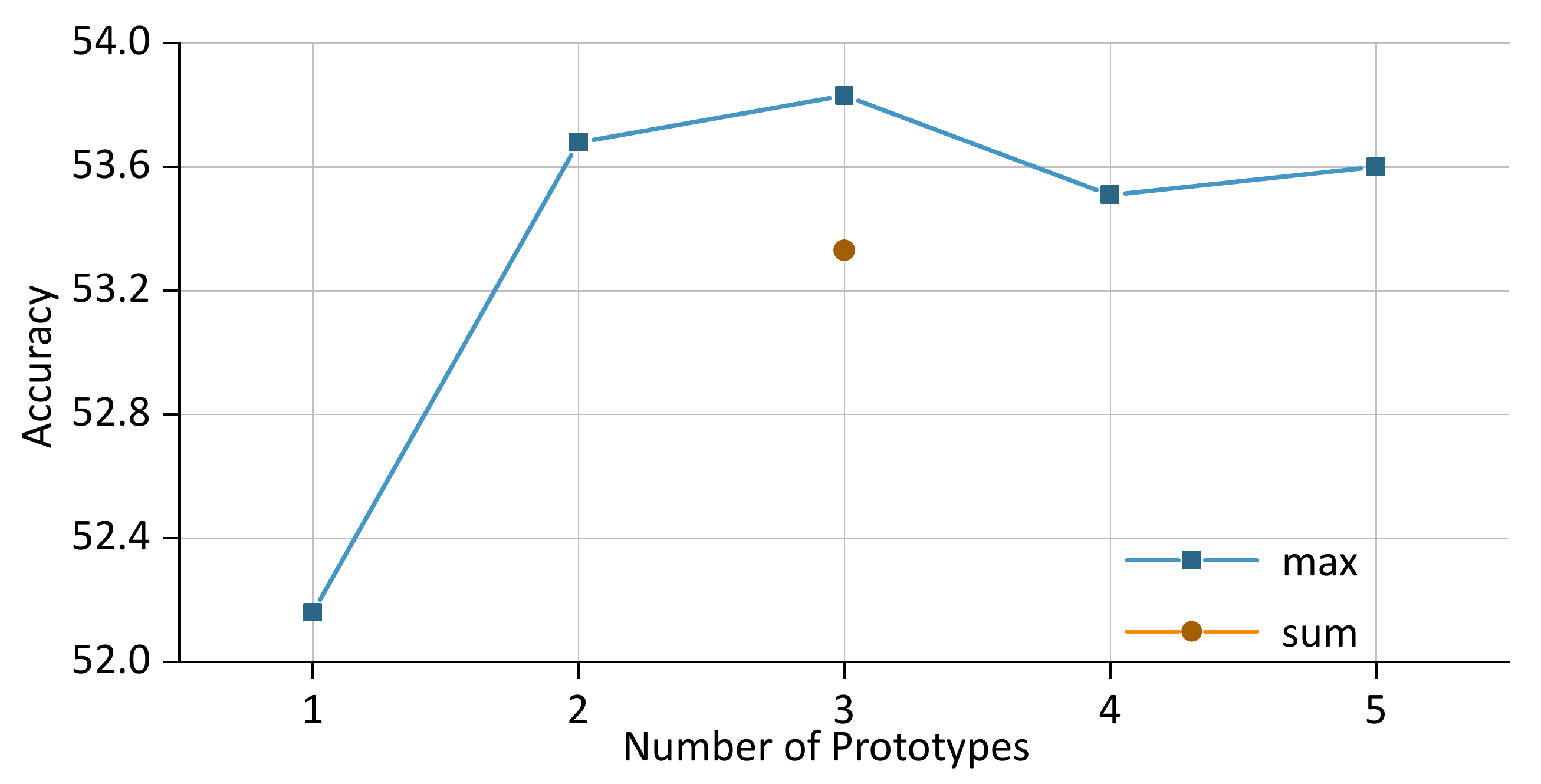}
\end{center}
  \caption{Ablation experiments of the number of meta-prototypes and the fusion strategy.}
\label{fig:number-prototypes}
\end{figure}

\paragraph{Meta-prototypes} In Fig.~\ref{fig:number-prototypes}, we conduct experiments to compare the performance with different meta-prototype numbers $M$ and the performance with an alternative fusion strategy---the {\it sum} operation for merging the predictions from multiple prototypes. The experiments are conducted in the prior network, and the number of foreground prototypes and background prototypes are changed simultaneously. One can observe that the model reaches the highest score when $M=3$. Using only one meta-prototype gives the worse result as the hierarchical information cannot be properly represented. Besides, excessive prototypes have no further contribution. 

Yang \etal.~\cite{yangPrototypeMixtureModels2020} merge multiple probability maps by directly summing them up. We explore the fusion strategy of multiple prototypes in Fig.~\ref{fig:number-prototypes}. One can observe that the {\it max} operation has better performance than that of the {\it sum} operation. We conjecture that the softmax in MPM forces the model to push multiple meta-prototypes encoding discrepant information. Merging multiple predictions by the {\it max} operation better conforms to the design of meta-prototypes. 

\begin{table}[!t]
\centering
\begin{tabular}{l|cccccc}
\toprule
Model & PN & SN & MP & CM & Accuracy \\
\midrule
Baseline &&&&& 50.03 \\ \hline
\multirow{2}{*}{1-stage} & \checkmark &&&& 52.06 \\
& \checkmark && \checkmark && 53.83 \\ \hline
\multirow{3}{*}{2-stage} & \checkmark & \checkmark &&& 53.75 \\ 
& \checkmark & \checkmark & \checkmark && 56.09 \\
& \checkmark & \checkmark & \checkmark & \checkmark & {\bf 56.52} \\
\bottomrule
\end{tabular}
\caption{Ablation experiments of the proposed modules. The mean-IoU results are reported in the 1-shot setting and the Resnet-50 feature extractor. PN: Prior Network; SN: Segmentation Network; MP: Meta-Prototypes; CM: Communication Modules.} \label{tab:structure}
\end{table}

\paragraph{Complexity analysis} 

We compare the model complexity of PEMP with previous methods in Table~\ref{tab:params}. Although PEMP is a two-stage model, the number of parameters does not increase too much comparing with one-stage methods, such as CaNet and RPMMs, and is even fewer than PFENet. The parameter number of the prior network is also listed for reference. To demonstrate that enlarging a model has no extra help for the performance, we fine-tune the prior network with a Resnet-101 feature extractor with more parameters than PEMP. The results suggest that it gets worse than PEMP or the prior network using Resnet-50, which indicates the importance of adequately leveraging the supervised knowledge in feature extraction.

The last column of Table~\ref{tab:params} shows the inference speed of different methods. We unify the input size of all the methods to $401\times 401$ for a fair comparison. PEMP is computationally efficient with the highest performance, achieving a real-time inference speed of 26 FPS. The prior network~(Resnet-50) and CaNet have a higher inference speed of 49 FPS and 42 FPS, but their performance is limited. The prior network~(Resnet-101) and PFENet have more parameters than PEMP, and their speeds are slower with 22 FPS. Although RPMMs have fewer parameters than PEMP, it embeds a mixture model and multiple residual branches in the network, slowing down the inference speed.

\begin{table}[!t]
\centering
\begin{tabular}{c|c|ccc}
\toprule
Model & Resnet & \#Param. & Accuracy & Speed \\
\midrule
Prior network & \multirow{5}{*}{50} & 12.0M & 53.83 & 49 \\
CaNet~\cite{Zhang_2019_CVPR} & & 19.0M & 54.46 & 42 \\
PFENet~\cite{tianPriorGuidedFeature2020} & & 34.4M & 55.98 & 22 \\
RPMMs~\cite{yangPrototypeMixtureModels2020} & & 19.6M & 56.10 & 23 \\
PEMP (Ours) & & 23.9M & 56.52 & 26 \\  \hline
Prior network & 101 & 27.5M & 53.77 & 22 \\
\bottomrule
\end{tabular}
\caption{Model complexity and inference speed comparison with state-of-the-arts. The prior network is a one-stage model. The unit of the speed is frame per second.} \label{tab:params}
\end{table}

\begin{table}[t]
\centering
\begin{tabular}{c|cc}
\toprule
Method & Accuracy \\
\midrule
w/o weight map& 52.63 \\
with weight map& 53.83 \\
\bottomrule
\end{tabular}
\caption{Results comparison when using or omitting the weight map in the model training. The experiments are conducted with the prior network in the 1-shot setting.} \label{tab:loss}
\end{table}

\begin{figure}[t]
\begin{center}
  \includegraphics[width=\linewidth]{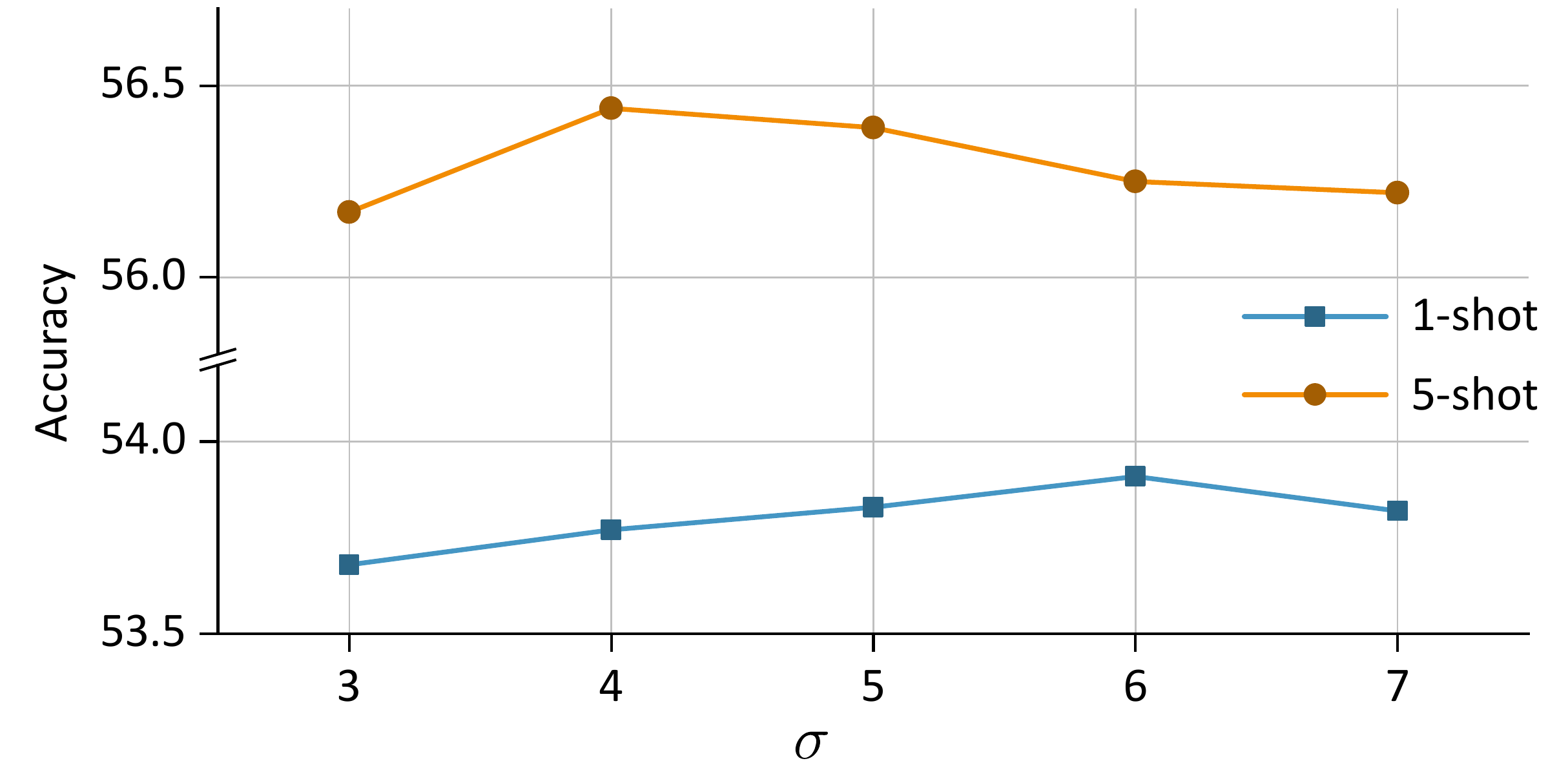}
\end{center}
  \caption{Comparison of different $\sigma$ values that are used in the weight map of the loss function.}
\label{fig:losses}
\end{figure}

\paragraph{Loss weights}

Table.~\ref{tab:loss} displays the segmentation results of using or omitting the weight map in the model training. It indicates that increasing the weights of boundary pixels helps the model improve segmentation accuracy. Fig.~\ref{fig:losses} shows the results using different $\sigma$ values, which are described in Eqn.~\eqref{eq:weight}. One can observe that the best performances of 1-shot and 5-shot are obtained at $\sigma=4$ and $\sigma=6$, respectively. Therefore, we take the average value of $\sigma=5$ by default.

\section{Conclusion}

We propose PEMP for few-shot segmentation based on learning more accurate features. The prior-enhanced network leverages the supervised knowledge in feature extraction, which guides the model to focus on the task-related objects and suppress noise information.
With multiple meta-prototypes, the model aggregates prior knowledge from the base classes to enhance the prototypes for novel classes and encodes hierarchical features to highlight the decision boundary. 
Experiments on PASCAL-$5^i$ and COCO-$20^i$ demonstrate that the proposed method improves the 5-shot results with a large margin comparing with either state-of-the-arts or the corresponding 1-shot results.

{\small
\bibliographystyle{ieee_fullname}
\bibliography{egbib}
}

\newpage
\section{Supplementary Material}

We present more details of the proposed Prior-Enhanced network with Meta-Prototypes~(PEMP) in Sec.~\ref{sec:structure}. We conduct extra ablation experiments about prior map transformation strategy and show the results in Sec.~\ref{sec:ablation}. Extra qualitative results of PASCAL-$5^i$ and COCO-$20^i$ are shown in Sec.~\ref{sec:result}. 

\subsection{Details of the Purifier}\label{sec:structure}

We adopt modified VGG-16 and Resnet-50 as the feature extractor of the prior-enhanced network. Following previous work~\cite{Zhang_2019_CVPR, Wang_2019_ICCV}, the last three convolutional layers of VGG-16 are replaced by dilated convolutional layers with a dilation rate of 2. For Resnet-50, only the first three residual blocks are reserved. Dilated convolution is applied to all the $3\times3$ convolutional layers in the third block. Different from CaNet~\cite{Zhang_2019_CVPR}, we do not fuse the multi-scale features of the second block and the third block, as no obvious improvement is observed in the experiments of the proposed method. 

\begin{figure}[h]
\begin{center}
  \includegraphics[width=\linewidth]{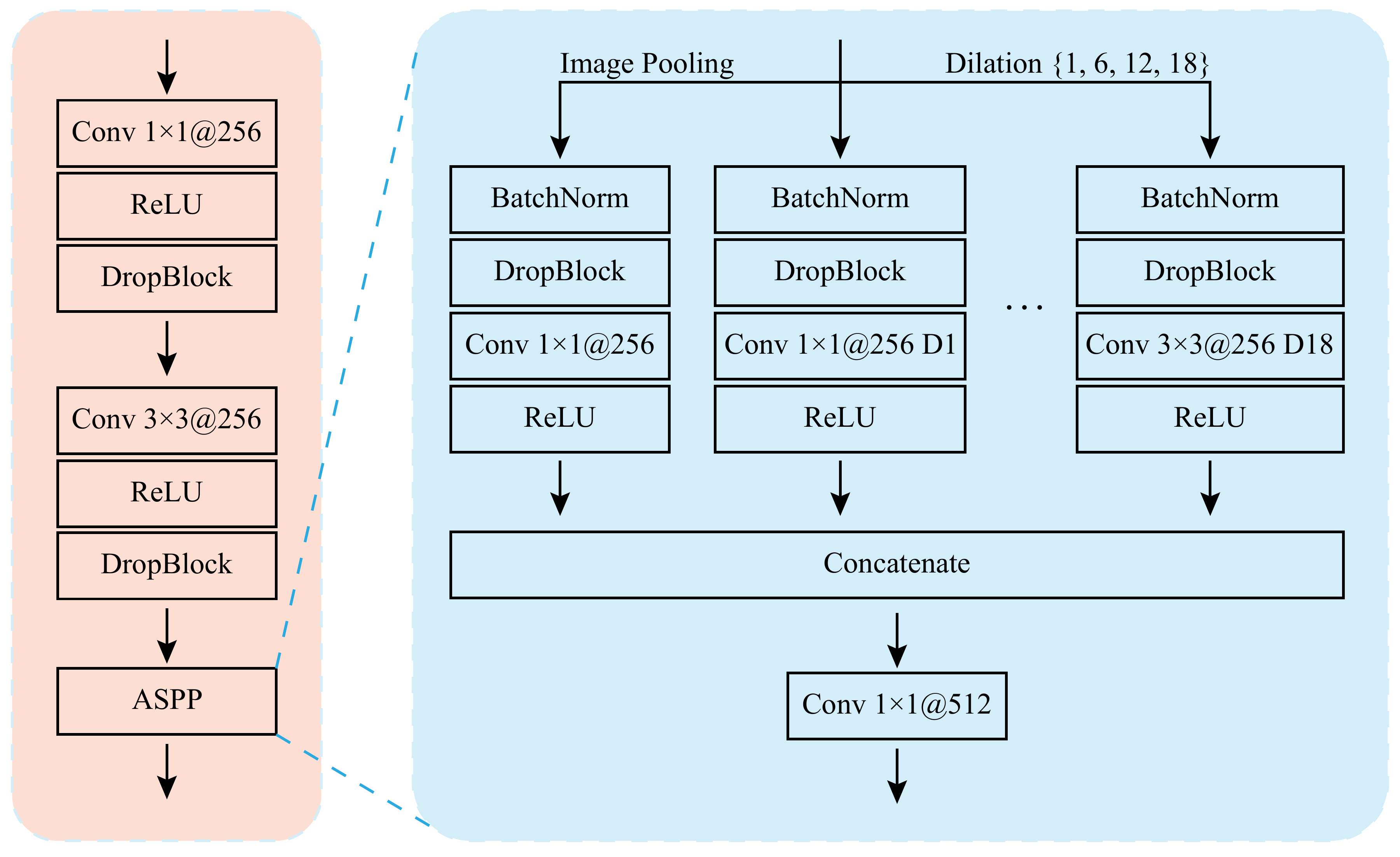}
\end{center}
  \caption{Detailed structure of the purifier in the prior network.} \label{fig:head-module-1}
\end{figure}

\begin{figure}[h]
\begin{center}
  \includegraphics[width=\linewidth]{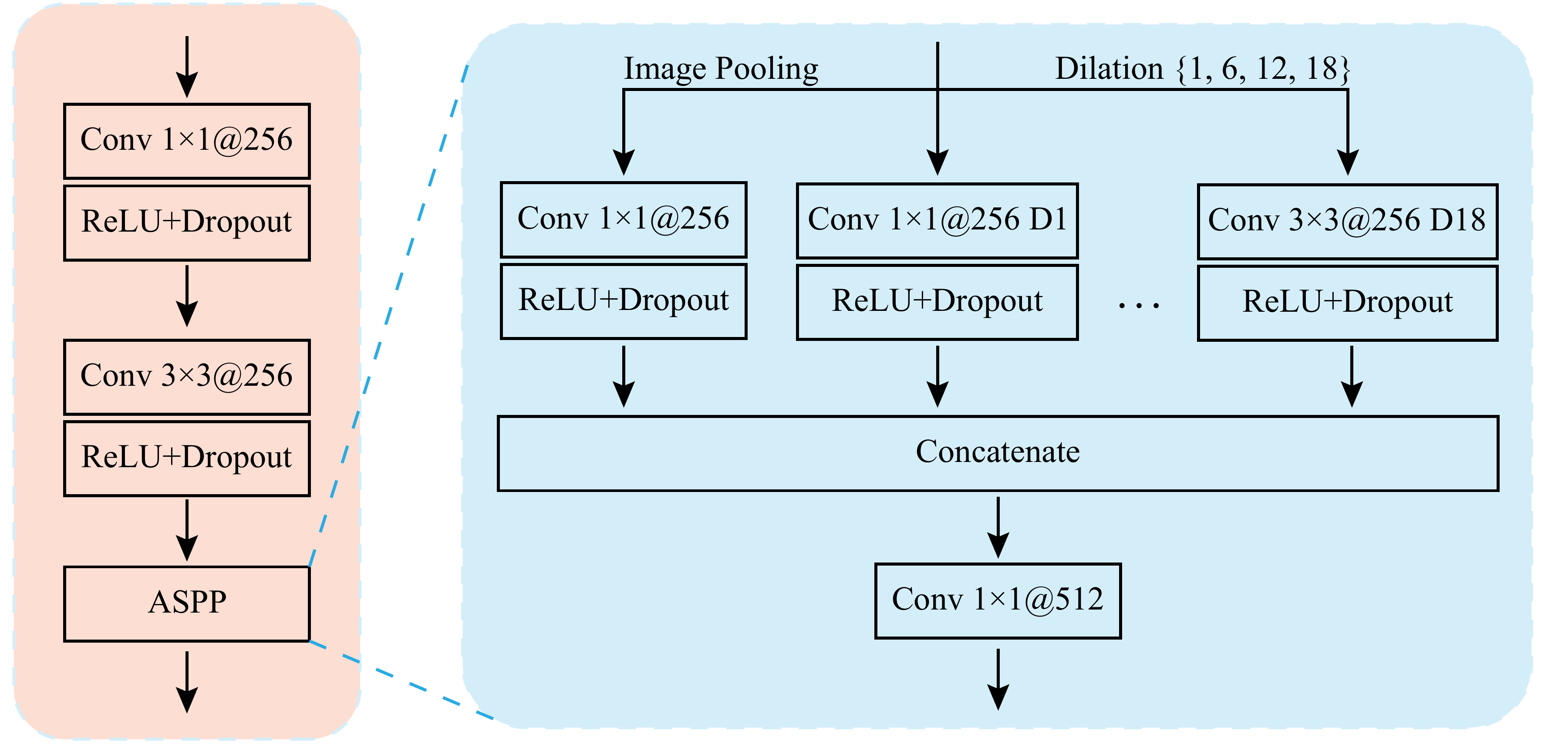}
\end{center}
  \caption{Detailed structure of the purifier in the segmentation network.} \label{fig:head-module-2}
\end{figure}

For Resnet, since residual paths bring low-level features into high-level layers, the feature extractor's outputs contain much noise. We introduce a purifier on top of the feature extractor to improve the smoothness of features. The purifier consists of a $1\times1$ convolutional layer used to reduce the number of channels, a $3\times3$ convolutional layer for smoothing the features, and an Atrous Spatial Pyramid Pooling (ASPP)~\cite{chenEncoderDecoderAtrousSeparable2018} for exploring multi-scale information. Details of the purifier are shown in Fig.~\ref{fig:head-module-1} and Fig.~\ref{fig:head-module-2}. We implement the prior network's purifiers and the segmentation network with sightly different structures to alleviate the overfitting problem. Concretely, for the prior network, the DropBlock layer~\cite{NEURIPS2018_7edcfb2d} randomly drop image blocks of size $4\times 4$ with a drop rate of 0.1. For the segmentation network, the Dropout layer has a drop rate of 0.5 for a stronger regularization. For VGG-16, we do not add the extra purifier for reducing the number of parameters, which still gives a good performance.

\subsection{Ablation Study}\label{sec:ablation}

\begin{table}[h]
\centering
\begin{tabular}{c|ccccc}
\toprule
Strategy & split-0 & split-1 & split-2 & split-3 & Mean \\
\midrule
Identity & 54.90 & 65.65 & {\bf 54.82} & 48.41 & 55.95 \\
Binary & {\bf 55.74} & 65.88 & 54.12 & {\bf 50.34} & {\bf 56.52} \\
\bottomrule
\end{tabular}
\caption{Mean-IoU of PEMP on PASCAL-$5^i$ with different strategies that how the query prior map is integrated into the segmentation network. Identity: the segmentation network directly takes the prior map as the pseudo-label of the query image; Binary: The prior map is binarized with a threshold of 0.5 and then treated as the pseudo-label.} \label{tab:strategy}
\end{table}

\paragraph{Prior map transformation strategy} The prior network produces a two-channel prior map which is the output of Softmax. We compare two strategies that how the prior map is transformed into the pseudo-label. 
\begin{itemize}
    \item {\bf Identity} The prior map is directly copied to the input layer as a pseudo-label. The range of prior map values is $[0, 1]$, while the range of the support label values is $\{0, 1\}$.
    \item {\bf Binary} The prior map is binarized with a threshould of 0.5, and threfore ``1'' denotes foreground pixels and ``0'' denotes background pixels. Then, the pseudo-label has the same meaning with support labels.
\end{itemize}

Table.~\ref{tab:strategy} shows the mean-IoU of two strategies in the 1-shot setting. Binary strategy gives the best performance on average. The Identity strategy has a lower score due to the misalignment of ranges between the support label and the pseudo-label. Although the binary pseudo-label has errors comparing with the real label, we treat the errors as data augmentation, which plays a role of regularizer. This regularization help the model avoid overfitting to the pseudo-labels but learn to fine-tune the pseudo-labels. 


\subsection{More qualitative results}\label{sec:result}

\begin{figure*}[h]
\begin{center}
  \includegraphics[width=\linewidth]{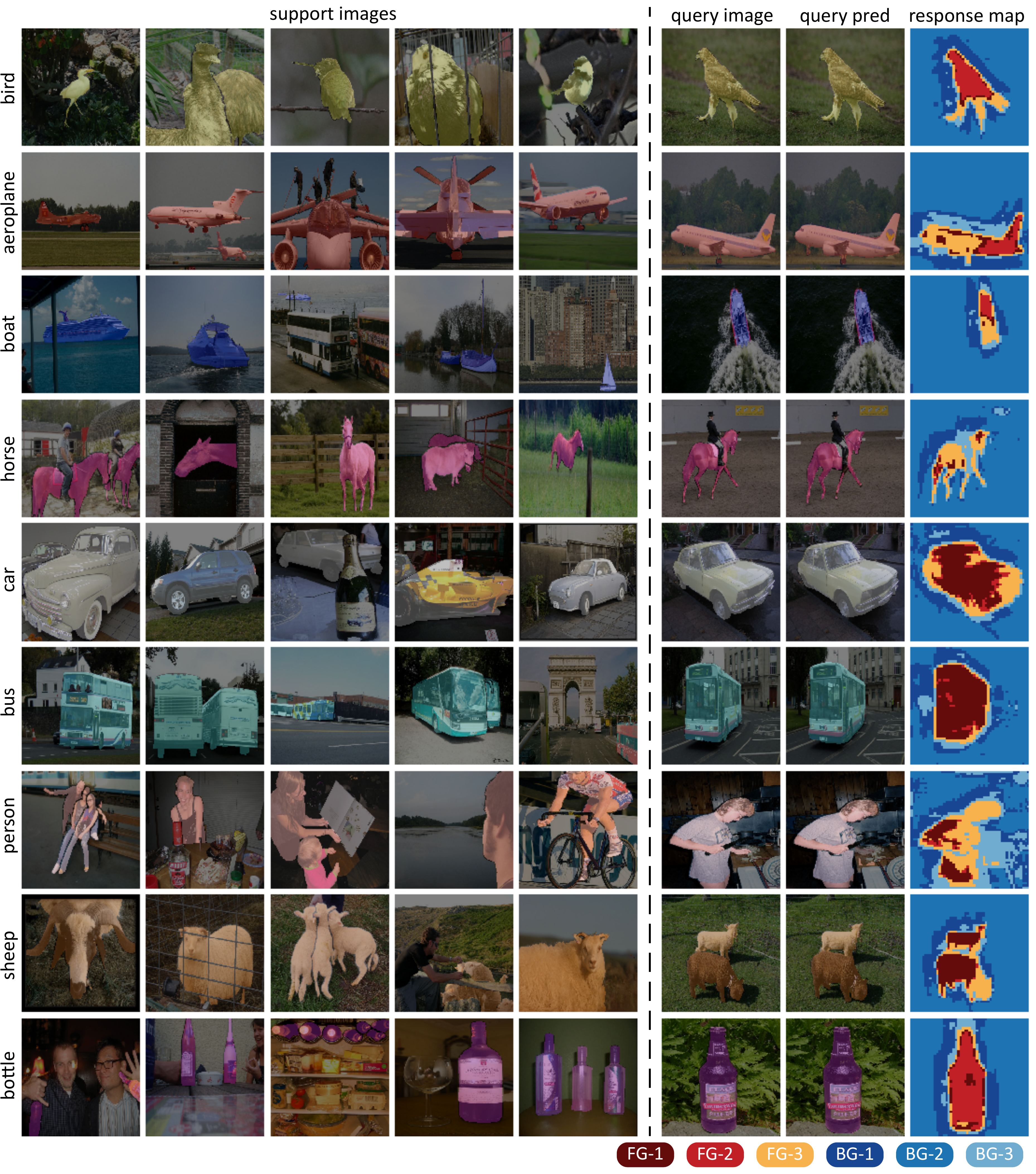}
\end{center}
  \caption{Qualitative results of our method in the 5-shot setting on PASCAL-$5^i$.}\label{fig:pascal-5-shot}
\end{figure*}

\begin{figure*}[h]
\begin{center}
  \includegraphics[width=\linewidth]{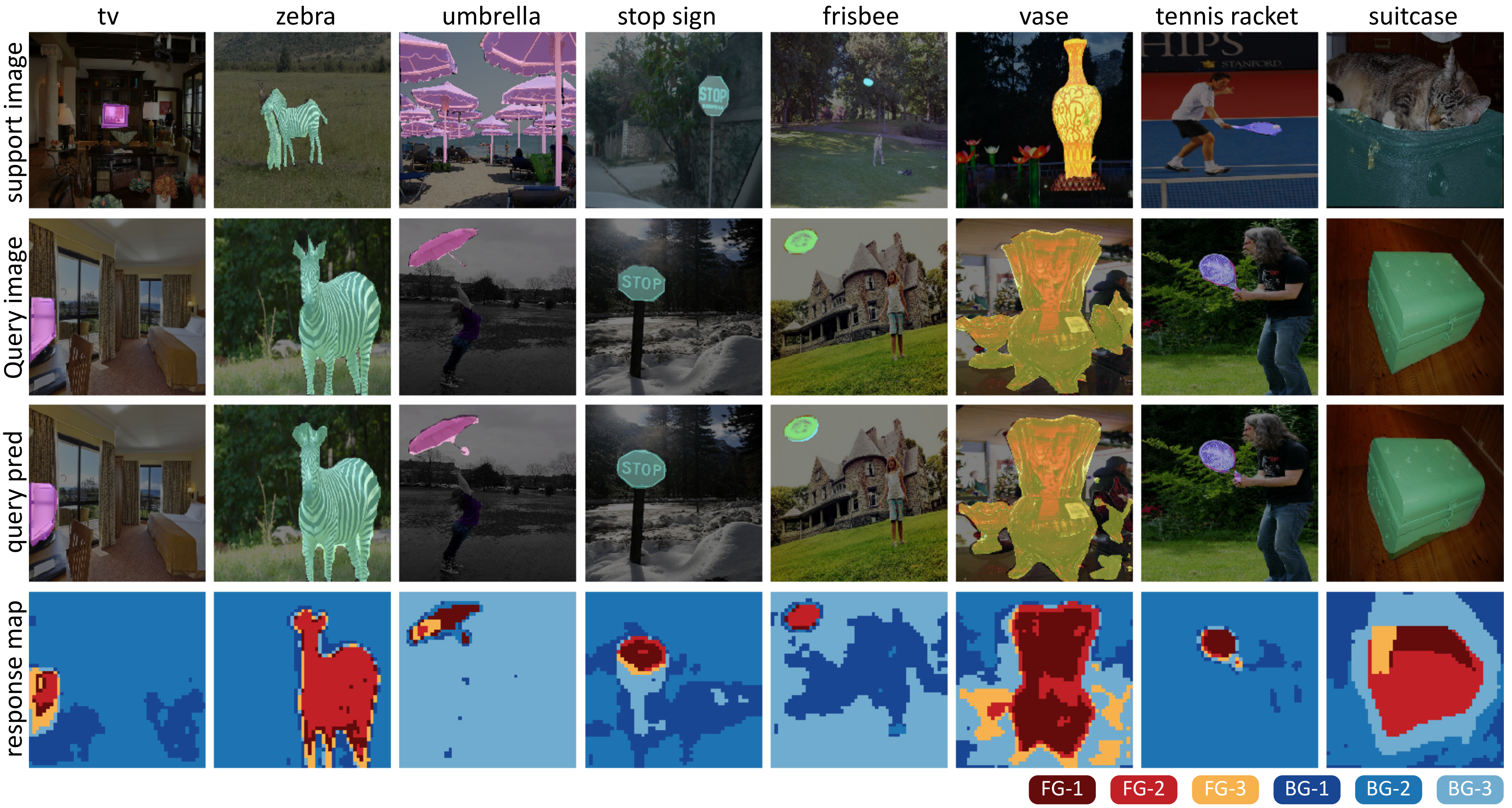}
\end{center}
  \caption{Qualitative results of our method in the 1-shot setting on COCO-$20^i$.}\label{fig:coco-1-shot}
\end{figure*}

Fig.~\ref{fig:pascal-5-shot} shows the results in the 5-shot setting on PASCAL-$5^i$. 
Fig.~\ref{fig:coco-1-shot} shows the results in the 1-shot setting on COCO-$5^i$. 

\end{document}